%% file: main.tex
\documentclass[10pt,twocolumn,letterpaper]{article}

\usepackage{dicta}
\usepackage{times}
\usepackage{epsfig}
\usepackage{graphicx}
\setkeys{Gin}{draft}
\usepackage{amsmath}
\usepackage{amssymb}
\usepackage{balance}

\usepackage[T1]{fontenc}
\usepackage{booktabs}
\usepackage{multirow}
\usepackage[table,dvipsnames]{xcolor}
\definecolor{intraavg}{RGB}{235,242,250}
\definecolor{interblock}{RGB}{239,248,242}
\definecolor{interavg}{RGB}{225,240,229}
\definecolor{oursgreen}{RGB}{242,248,242}
\usepackage{tikz}
\usepackage{amsfonts}
\usepackage{rotating}
\usepackage{comment}
\usepackage{subcaption}
\usepackage{array}
\usepackage{cite}
\usepackage{IEEEtrantools}
\usepackage{ragged2e}

\newcommand{\coolname}{DAD}

\usepackage[pagebackref=true,breaklinks=true,letterpaper=true,colorlinks,bookmarks=false]{hyperref}

\dictafinalcopy %

\def\dictaPaperID{99} %

\ifdictafinal\fi

\newlength{\copyrightleft}
\begin{document}

\bstctlcite{IEEEexample:BSTcontrol}

\title{Geometry-Conditioned Visual Place Recognition in Natural Environments}

\author{
Walter Nedov$^{1}$, Saimunur Rahman$^{1}$, Kavindie Katuwandeniya$^{1}$, \\ David Hall$^{1}$, Kaushik Roy$^{1}$,  Peyman Moghadam$^{1,2}$ \\
{\small $^{1}$CSIRO Robotics, CSIRO, Australia \quad
\texttt{firstname.lastname@csiro.au}}\\
{\small $^{2}$Queensland University of Technology, Australia \quad
\texttt{firstname.lastname@qut.edu.au}}
}

\maketitle

\begin{tikzpicture}[remember picture,overlay]
\node[
    anchor=south west,
    text width=\textwidth,
    inner xsep=0pt,
    inner ysep=0pt,
    align=center,
    font=\fontsize{7.5}{8.2}\selectfont
] at ([xshift=\copyrightleft,yshift=0.25in]current page.south west) {%
    \noindent\justifying
    \textcopyright~2026 IEEE. Personal use of this material is permitted.
    Permission from IEEE must be obtained for all other uses, in any current
    or future media, including reprinting/republishing this material for
    advertising or promotional purposes, creating new collective works,
    for resale or redistribution to servers or lists, or reuse of any
    copyrighted component of this work in other works.
};
\end{tikzpicture}

\input{sections/abstract}

\input{sections/introduction}

\input{sections/related_works}

\input{sections/methodology}

\input{sections/implementation}

\input{sections/experiments}
\input{sections/conclusion}

\balance{}

{\small
\bibliographystyle{IEEEtran}
\bibliography{references}
}

\end{document}

%% file: sections/abstract.tex
\begin{abstract}
Visual Place Recognition (VPR) in natural environments remains challenging due to repetitive vegetation, sparse distinctive landmarks, and substantial appearance and viewpoint variation across traversals. While visual observations of the same place can change considerably, their underlying spatial structure is often more persistent. We exploit this complementary geometric consistency through Depth-Aware Distillation (\coolname{}), which conditions the token representations of a pretrained Vision Foundation Model (VFM) on geometry inferred by a Geometric Foundation Model (GFM), without any depth sensor. Rather than treating geometry as an additional input modality, \coolname{} projects image-aligned depth into the VFM token space and selectively modulates visual representations through channel-wise geometric conditioning. A two-stage teacher-guided learning strategy first anchors the geometry-conditioned representation to the pretrained appearance space, before refining it for place discrimination. Evaluated on the WildCross benchmark, \coolname{} improves average inter-sequence Recall@1 from 61.41\% to 66.37\% and Recall@5 from 65.86\% to 72.49\% over a matched appearance-only baseline, with the largest gains under reverse traversal and long-term appearance variation. These results show that GFM-derived geometry can provide a persistent structural prior for VPR when visual appearance becomes unreliable.

\end{abstract}
\vspace{-4mm}

%% file: sections/introduction.tex
\section{Introduction}

\begin{figure}[t]
    \centering
    \includegraphics[draft=false,width=0.9\columnwidth]{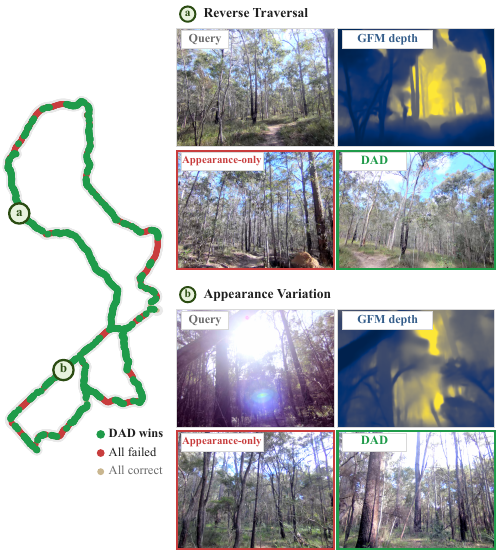}
    \caption{Geometry complements appearance-based VPR in natural environments. \coolname{} corrects appearance-only retrieval failures under reverse traversal and appearance variation by exploiting GFM-derived scene structure.}
    \label{fig:hero-figure}
    \vspace{-5.0mm}
\end{figure}

Place Recognition (PR) is the task of determining whether a current sensor observation corresponds to a previously visited place, making it a core capability for autonomous robotic systems operating in GPS-denied/limited environments such as underground spaces, dense forests, and indoor environments~\cite{yin2025general}. PR has been studied across complementary sensing modalities, including Visual~\cite{izquierdo_optimal_2024, hausler_pair-vpr_2025}, LiDAR~\cite{vidanapathirana_logg3d-net_2022, jung2025imlpr}, and Radar~\cite{kim2025sherloc}, each providing different appearance, geometric, and structural cues for localization.  However, reliable PR in natural and unstructured environments remains particularly challenging due to repetitive vegetation, sparse distinctive landmarks, irregular scene structure, and substantial changes in viewpoint and appearance across revisits~\cite{knights_wild-places_2023,knights_wildcross_2026}.

Visual Place Recognition (VPR), in particular, is strongly affected by these conditions because it relies on distinctive and repeatable visual cues. Current VPR methods increasingly build on Vision Foundation Models (VFMs), such as DINOv2, which provide strong semantic representations and achieve high performance in structured urban environments~\cite{izquierdo_optimal_2024, hausler_pair-vpr_2025}. In such environments, persistent buildings, road markings, signs, and other landmarks provide reliable semantic cues. In contrast, natural environments often contain extensive repetitive vegetation and fewer distinctive landmarks, resulting in perceptual aliasing between geographically different places. Changes in illumination, weather, season, vegetation growth, and viewpoint can further alter the appearance of the same place across revisits~\cite{knights_wildcross_2026, leyva_vallina_gardens_2019}, while reverse traversal can produce substantially different observations despite relatively stable scene structure. Fig.~\ref{fig:hero-figure} illustrates these two representative failure modes, where appearance-only retrieval fails under reverse traversal and severe appearance variation, while estimated depth preserves the coarse geometry of the path and surrounding vegetation. This motivates the use of geometry as a complementary signal for VPR. While LiDAR-based methods demonstrate the robustness of geometric structure, they require additional sensing hardware. Recent Geometric Foundation Models (GFMs), such as Depth Anything V2~\cite{yang_depth_2024}, can instead estimate dense scene geometry directly from RGB images, providing spatially aligned geometric cues that can be integrated with VFM representations. 

Motivated by these observations, we introduce \underline{D}epth-\underline{A}ware \underline{D}istillation (\coolname{}), a geometry-conditioned VPR framework that uses depth inferred by a Geometric Foundation Model (GFM) to guide the token representations of a pretrained Vision Foundation Model (VFM), without requiring a dedicated depth sensor. Rather than treating geometry as an additional input modality, \coolname{} projects image-aligned depth into the VFM token space and selectively modulates the visual representation through learnable channel-wise geometric conditioning. To preserve the pretrained appearance representation, \coolname{} adopts a two-stage teacher-guided training strategy that first anchors the geometry-conditioned descriptor to a frozen appearance-based teacher before refining it for place discrimination.

%% file: sections/related_works.tex
\section{Related Work}

\noindent \textbf{Visual Place Recognition:}
VPR is commonly formulated as image retrieval, in which a query image is matched against a database of images through a global descriptor. Classical pipelines aggregated hand-crafted local features using vocabulary-based encodings such as VLAD. Earlier work addressed appearance change through condition-invariant global-to-local image comparison~\cite{milford_condition_2014}. NetVLAD~\cite{arandjelovic_netvlad_2016} made feature aggregation differentiable, and subsequent work has increasingly adopted vision foundation models as feature extractors. DINOv2~\cite{oquab_dinov2_2024} produces semantically rich patch features that now underpin leading VPR methods. BoQ~\cite{ali-bey_boq_2024} pools features using learnable queries, SALAD~\cite{izquierdo_optimal_2024} formulates aggregation as optimal transport-based feature-to-cluster assignment, and Pair-VPR~\cite{hausler_pair-vpr_2025} employs masked image modeling to train a pair-classifier decoder that informs its encoder's retrieval objective. While these models excel in structured domains, their visual features do not readily generalize to unstructured natural scenes~\cite{keetha_anyloc_2023}, motivating the integration of geometric priors into the visual representation space.
\vspace{1mm}

\noindent \textbf{Geometric Foundation Models:}
Parallel to the development of VFMs for semantic understanding, Geometric Foundation Models (GFMs) have emerged that infer 3D scene structure directly from images. Monocular models such as Depth Anything V2~\cite{yang_depth_2024} predict dense depth maps from a single RGB frame, while multi-view models such as DUSt3R~\cite{wang_dust3r_2024}, MASt3R~\cite{leroy_mast3r_2024} VGGT~\cite{wang2025vggt} jointly estimate camera poses, depth, and dense point maps using a single feed-forward transformer. These GFMs produce geometric representations such as depth images and point cloud projections that generalize across domains without scene-specific training. Recent work has shown that such features can be transferred to downstream vision models to refine their representations, measurably improving spatial understanding in models that otherwise lack explicit 3D reasoning~\cite{yang_understanding_2026}. 
Our work takes inspiration from this finding to suggest that GFM features could similarly enrich the representations of appearance-only VFMs for VPR, providing geometric priors that can support retrieval in challenging environments.

\vspace{2mm}

\noindent \textbf{Integrating GFM Features into VFMs:}
Multimodal integration in pretrained foundation models requires a mechanism for incorporating additional signals without disrupting the original representation space. GFM-derived tokens are not naturally aligned with a VFM's feature space, so naive feature injection can disrupt the pretrained representation without improving its utility for retrieval. RGB-D architectures such as DFormer~\cite{yin_dformer_2024} learn dedicated multimodal backbones for scene understanding, but are not designed to preserve the descriptor space of an existing pretrained VPR model. Joint multimodal training can also suffer from the dominating-modality problem, in which one modality dominates the descriptor space~\cite{komorowski_minkloc_2021}. Alignment-based approaches instead align a new modality to a frozen foundation encoder before fusion, whether across large paired datasets (ViT-Lens~\cite{lei_vit-lens_2024}) or through token-level distillation over compatible patch tokens that avoids the cross-architecture gap of heterogeneous backbones (TOLiD ~\cite{tolid_2026}).
\coolname{} applies this alignment principle within a single place descriptor by using a frozen appearance teacher to anchor the geometry-conditioned descriptor to the pretrained VFM retrieval space, as detailed in the following section.

%% file: sections/methodology.tex
\section{Methodology}

\begin{figure*}[!th]
    \vspace{-1.0em}
    \centering
    \includegraphics[draft=false,width=\linewidth]{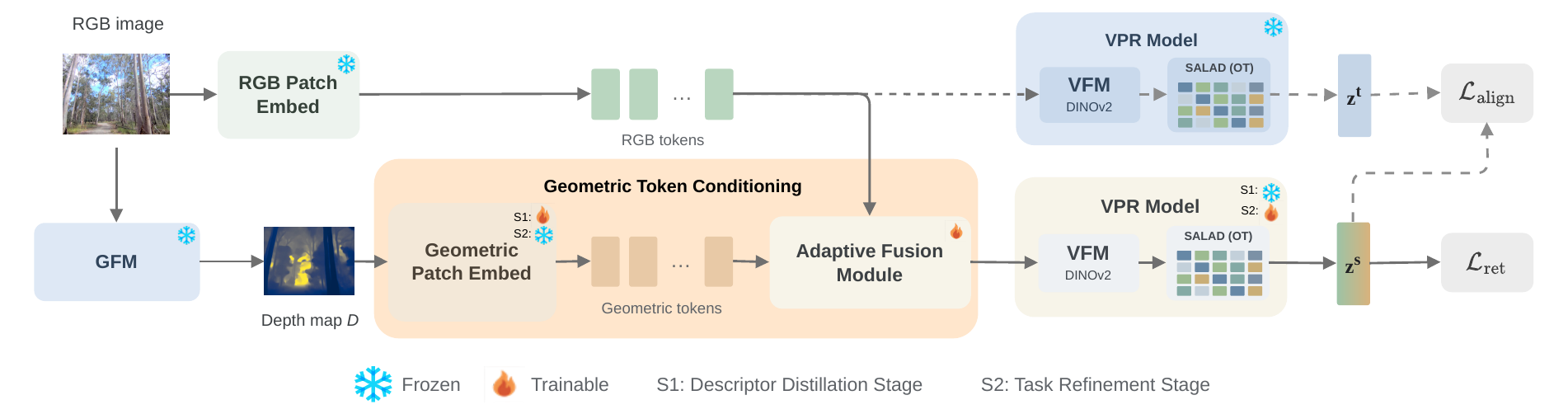}
    \vspace{-1.5em}
    \caption{Overview of \coolname{}. GFM-derived geometric tokens condition the VFM appearance representation to produce a global place descriptor. During descriptor distillation, an appearance-only teacher anchors the geometry-conditioned representation through $\mathcal{L}_{\mathrm{align}}$, together with the retrieval objective $\mathcal{L}_{\mathrm{ret}}$. During task refinement, the teacher constraint is removed and the representation is optimized for place discrimination using $\mathcal{L}_{\mathrm{ret}}$.}
    \label{fig:main-framework}
    \vspace{-1.0em}
\end{figure*}

\coolname{} conditions the representation of a pretrained appearance-based VPR model using geometry inferred by a pretrained GFM. Given an RGB image, the VFM produces appearance tokens, while the GFM estimates an image-aligned depth map that provides geometric information about the scene. The GFM remains fixed throughout training, whereas the VPR model is adapted in stages to incorporate this geometric information without disrupting its established appearance representation. As illustrated in Fig.~\ref{fig:main-framework}, \coolname{} projects the estimated depth into the VFM token space and selectively integrates it with the appearance tokens before producing a global place descriptor.

\coolname{} adopts a two-stage training strategy. During \emph{descriptor distillation}, the appearance-based VPR model is kept fixed and serves as a teacher, while the newly introduced geometry-conditioning components are learned. This stage establishes compatibility between the geometric information and the existing VPR representation. During \emph{task refinement}, selected higher-level layers of the VPR model are fine-tuned for place discrimination, while the learned geometric embedding is kept fixed.

\subsection{Geometry-Conditioned Token Embedding}
\label{sec:fusion}

\begin{figure}[b!]
    \centering
    \includegraphics[draft=false,width=0.8\linewidth]{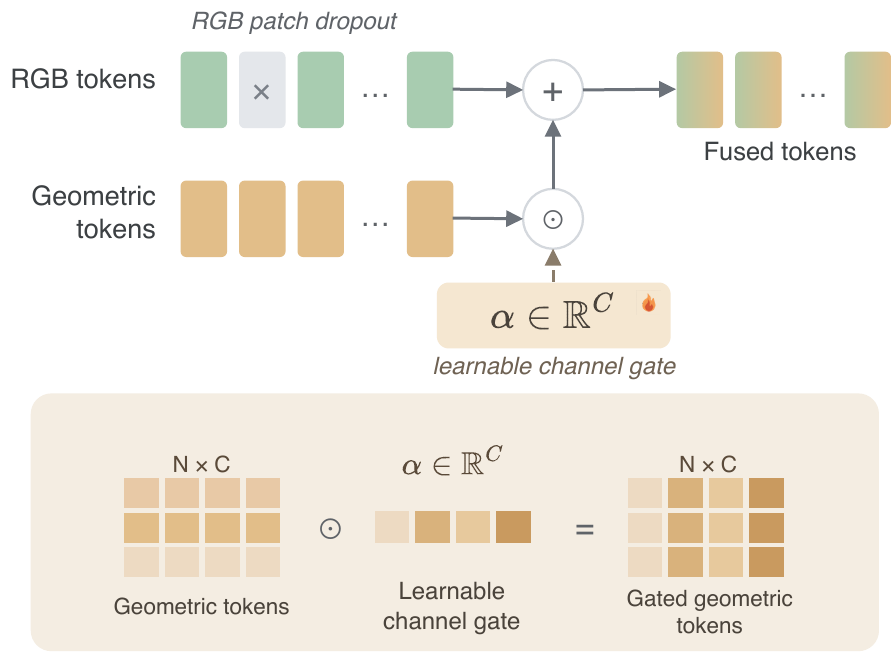}
\caption{Adaptive Fusion Module. GFM-derived geometric tokens are modulated by a learnable channel-wise gate and fused with the VFM tokens. RGB patch dropout is applied during training to regularize the appearance features.}
\label{fig:fusion}
\vspace{-4mm}
\end{figure}

\coolname{} introduces geometry directly into the token space of the appearance-based VFM while preserving the spatial correspondence between visual and geometric information. Let $E_v(\cdot)$ denote the patch embedding layer of the VFM, which maps an RGB image $I$ to a sequence of appearance tokens $E_v(I)$. Given the same image, the GFM produces an image-aligned depth map $D$.

To represent this depth information in the VFM token space, \coolname{} introduces a learnable geometric embedder $E_g(\cdot)$, implemented as a lightweight convolutional projection module. The embedder transforms $D$ into geometric tokens with the same spatial resolution and channel dimension as the appearance tokens:

\begin{equation}
    \mathbf{g} = E_g(D) \in \mathbb{R}^{N \times C},
\end{equation}
where $N$ denotes the number of spatial patches and $C$ is the VFM token dimension.
Although appearance and geometry provide complementary information, directly adding geometric tokens to the appearance representation can disrupt the feature space learned by the VFM. Moreover, the usefulness of geometric information may vary across feature channels. To selectively regulate its contribution, we introduce an \emph{Adaptive Fusion Module} that applies a learnable channel-wise gate:

\begin{equation}
\mathbf{x} = E_v(I) + \boldsymbol{\alpha} \odot E_g(D),
\label{eq:fusion}
\end{equation}
where $\boldsymbol{\alpha}\in\mathbb{R}^{C}$ is a learnable channel-wise gating vector and $\odot$ denotes channel-wise multiplication broadcast across all spatial tokens. %
As illustrated in Fig.~\ref{fig:fusion}, the same gate is shared across spatial tokens, while each feature channel can independently regulate the contribution of geometry. The gate is initialized near zero, with $\alpha = 0.01$ for all channels, so that the initial geometry-conditioned representation remains close to the original appearance representation. During training, the gate learns which channels benefit from geometric conditioning while allowing appearance-dominant channels to remain largely unchanged.

The resulting geometry-conditioned token sequence $\mathbf{x}$ is then processed by the VFM transformer blocks and the VPR aggregation head. Let $T_v(\cdot)$ denote the VFM transformer blocks and $G(\cdot)$ the global descriptor aggregation head, instantiated using SALAD in our experiments. The resulting global place descriptor $\mathbf{z}$ is given by

\begin{equation}
    \mathbf{z} = G(T_v(\mathbf{x})).
\label{eq:fused_descriptor}
\end{equation}

This token-level conditioning has two advantages. First, image-aligned depth preserves spatial correspondence between appearance and geometric tokens, allowing the transformer to jointly reason over both cues. Second, geometry is introduced directly into the existing VFM representation rather than through a separate geometry-specific backbone, retaining the original VPR architecture with only the geometric embedder and channel-wise gate added.

\subsection{Descriptor Distillation Stage}
\label{sec:stage1}

The geometry-conditioned representation should introduce complementary structural information without losing the discriminative aspects of the appearance-based descriptor space. To mitigate this, \coolname{} first performs a descriptor distillation stage that learns to incorporate geometric information while maintaining consistency with the established appearance representation.

We formulate this stage using a teacher-student framework. The appearance-only VPR model is kept fixed and serves as the teacher, providing a stable reference descriptor space, while the geometry-conditioned pathway forms the student. Given an RGB image $I$, the teacher descriptor is

\begin{equation}
\mathbf{z}^{t} = G(T_v(E_v(I))).
\end{equation}

While the student descriptor $\mathbf{z}^{s}$ is obtained from the geometry-conditioned representation defined in Eqs.~\ref{eq:fusion}--\ref{eq:fused_descriptor}. During descriptor distillation, the VFM backbone and aggregation head are kept fixed, and only the geometric embedder $E_g(\cdot)$ and channel-wise gate $\boldsymbol{\alpha}$ are optimized.

The descriptor distillation stage is optimized with two complementary objectives. First, the geometry-conditioned representation should remain discriminative for place retrieval. We therefore employ the multi-similarity retrieval loss~\cite{wang_multi-similarity_2020}:

\begin{equation}
\begin{split}
\mathcal{L}_{\mathrm{ret}} = \frac{1}{|\mathcal{B}|} \sum_{i \in \mathcal{B}} \bigg[ &\frac{1}{\kappa_p} \log\left(1 + \sum_{k \in \mathcal{P}_i} e^{-\kappa_p(S_{ik} - \gamma)}\right) \\
&+ \frac{1}{\kappa_n} \log\left(1 + \sum_{k \in \mathcal{N}_i} e^{\kappa_n(S_{ik} - \gamma)}\right) \bigg],
\end{split}
\label{eq:retrieval_loss}
\end{equation}
where $S_{ik}$ denotes the cosine similarity between the descriptors of samples $i$ and $k$, and $\mathcal{P}_i$ and $\mathcal{N}_i$ denote the sets of positive and negative samples associated with anchor $i$, respectively.  
Second, to constrain the geometry-conditioned descriptor from drifting excessively from the established appearance-based representation, we introduce a descriptor-level alignment loss between the student and teacher:

\vspace{-3mm}
\begin{equation}
\mathcal{L}_{\mathrm{align}}
=
1-\frac{\langle \mathbf{z}^{s},\mathbf{z}^{t}\rangle}
{\|\mathbf{z}^{s}\|_2 \, \|\mathbf{z}^{t}\|_2},
\label{eq:align_loss}
\end{equation}
where $\mathbf{z}^{s}$ and $\mathbf{z}^{t}$ denote the student and teacher descriptors, respectively. The cosine alignment objective preserves consistency with the teacher descriptor space while still allowing the geometry-conditioned representation to introduce complementary information for place discrimination.
The retrieval and alignment losses jointly define the objective for the descriptor distillation stage ($\mathcal{L}_{\mathrm{distill}}$):

\begin{equation}
    \mathcal{L}_{\mathrm{distill}}
    =
    \mathcal{L}_{ret}
    +
    \lambda \mathcal{L}_{\mathrm{align}},
    \label{eq:total_loss}
\end{equation}
where $\lambda$ controls the contribution of the teacher-alignment objective. 
The retrieval loss encourages the geometry-conditioned representation to exploit complementary geometric information for improved place discrimination, while the alignment loss limits excessive deviation from the appearance-based teacher representation. Together, these objectives enable \coolname{} to learn a geometry-conditioned descriptor that preserves the established retrieval structure while incorporating additional geometric cues.

\subsection{Task Refinement Stage}
\label{sec:stage2}

Descriptor distillation optimizes
$\mathcal{L}_{\mathrm{distill}}$ to introduce geometric information while
constraining the representation to remain consistent with the
appearance-based teacher. However, this alignment constraint can limit the extent to which the representation adapts to the discriminative requirements of VPR. We therefore introduce a task refinement stage that relaxes this constraint and optimizes the geometry-conditioned representation directly for place discrimination.

During this stage, the appearance-only teacher is no longer used and the geometry-conditioned pathway is optimized using only the retrieval objective. The geometric embedder $E_g(\cdot)$ is kept fixed to preserve the depth-to-token mapping learned during descriptor distillation. In contrast, the higher-level components of the VPR model, including the final two VFM transformer blocks and the SALAD aggregation head, are fine-tuned to adapt the joint appearance-geometry representation to the target retrieval task. The channel-wise gate $\boldsymbol{\alpha}$ remains trainable, allowing the contribution of geometric information to be recalibrated as the representation becomes more discriminative for VPR.
The task refinement objective is therefore

\begin{equation}
    \mathcal{L}_{\mathrm{refine}}
    =
    \mathcal{L}_{ret}.
    \label{eq:refinement_loss}
\end{equation}

By separating representation alignment from task-specific adaptation, the two training stages serve complementary roles. Descriptor distillation introduces geometric information while maintaining consistency with the established appearance-based descriptor space, whereas task refinement subsequently adapts the geometry-conditioned representation for improved place discrimination.

%% file: sections/implementation.tex
\section{Experimental Setup}

\subsection{Dataset and GFM Adaptation}

We evaluate \coolname{} on the WildCross benchmark~\cite{knights_wildcross_2026}, which provides over 476{,}000 sequential RGB frames collected across two natural forests in southeast Queensland, Australia: Karawatha and Venman. Within each location there are four traversal sequences. Sequence 01 is the reference forward traversal, sequence 02 retraces the route in reverse on the same day, sequence 03 follows an alternate extended path, and sequence 04 repeats the reference trajectory after 14 months providing vegetational change due to the passage of time. All frames are paired with GPS/IMU poses and synchronized LiDAR submaps.

We adopt Depth Anything V2 (DA2)~\cite{yang_depth_2024} with a ViT-B encoder as our GFM. We adapt DA2 to the forest domain using sparse WildCross LiDAR projections as metric-depth supervision, following the depth-adaptation protocol and hyperparameters of WildCross~\cite{knights_wildcross_2026}. This adaptation is performed offline, after which the GFM remains frozen and \coolname{} operates on RGB imagery and GFM-estimated depth without requiring LiDAR. 
Using the adapted checkpoint, we generate dense depth maps at DA2's native 518-pixel input resolution. Before being provided to \coolname{}, each depth map is resized alongside its corresponding RGB frame and normalized per image to zero mean and unit variance.

\subsection{Training Details}

Following the WildCross training implementation~\cite{knights_wildcross_2026}, positive pairs are defined by a GPS distance within $5\,\mathrm{m}$ and heading alignment within $\pm10^{\circ}$, while negatives must lie beyond $50\,\mathrm{m}$. Each batch contains 128 images sampled from 32 places, with four images per place. Images are resized to $224 \times 224$ for training and $322 \times 322$ for evaluation.

All augmentations are applied only during training, sampled independently for each image. Horizontal flips are applied jointly to the RGB image and depth map with probability 0.5 to preserve their spatial alignment. We additionally apply RGB-only Gaussian blur with probability 0.3. RGB patch dropout is applied with probability 0.3 by randomly masking 25\% of the appearance tokens.

\coolname{} is trained for 40 epochs in two 20-epoch stages using AdamW with weight decay $10^{-4}$. In the first stage, only the geometric embedder $E_g$ and gate $\boldsymbol{\alpha}$ are trained at a learning rate of $10^{-4}$, using a 5-epoch linear warmup followed by MultiStepLR decay and an alignment weight of $\lambda=0.05$. The second stage initialises from the first and keeps $E_g$ frozen while fine-tuning the final two DINOv2 transformer blocks and SALAD aggregation head at $10^{-5}$, with $\boldsymbol{\alpha}$ remaining trainable at $10^{-4}$. This stage uses a 2-epoch warmup with MultiStepLR decay at epoch 15.

\subsection{Evaluation Protocol}

Following the WildCross protocol~\cite{knights_wildcross_2026}, each split holds out one sequence from each location for evaluation, while the remaining three sequences per location are used for training. The held-out sequences serve as query sets. For inter-sequence retrieval, the remaining sequences from the same location form the database. For intra-sequence retrieval, the held-out sequence serves as both the query and database, with a 600\,s temporal exclusion window. Database images are ranked by cosine similarity between global descriptors, and a retrieval is considered correct if it lies within 25\,m of the query. We report Recall@1 and Recall@5 averaged over the four splits. Baselines include a matched appearance-only SALAD model, NetVLAD~\cite{arandjelovic_netvlad_2016}, and MixVPR~\cite{ali-bey_mixvpr_2023}.

%% file: sections/experiments.tex
\section{Results and Discussion}
\input{figures/DAD_RESULTS}

\paragraph*{Retrieval Performance}

The main quantitative results are reported in Tables~\ref{tab:dad_intra_results} and~\ref{tab:dad_inter_results}. \coolname{} achieves the best average performance across both evaluation settings, improving intra-sequence Recall@1 from 61.76 to 65.34. However, the inter-sequence results best demonstrate \coolname{}'s strength in providing robust place recognition across visually ambiguous and inconsistent observations, with Recall@1 increasing from 61.41 to 66.37 and Recall@5 from 65.86 to 72.49 over the matched appearance-only SALAD control.

\begin{figure}[h]
    \vspace{-0.5em}
    \centering
    \includegraphics[draft=false,width=\columnwidth]{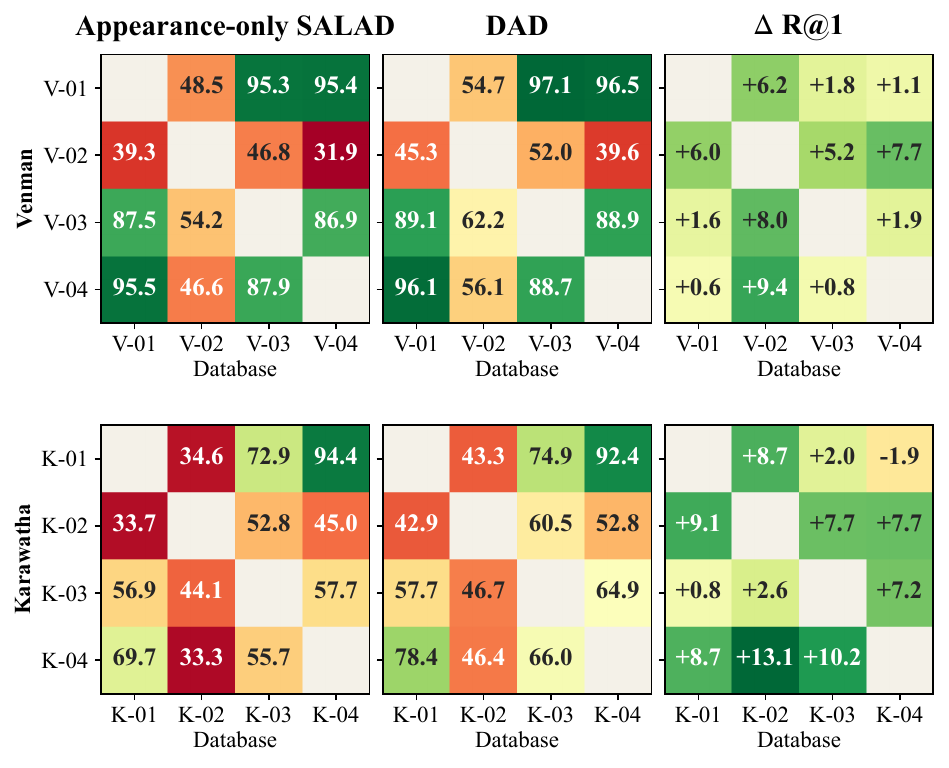}
    \caption{Inter-sequence Recall@1 on WildCross for the matched SALAD control, \coolname{}, and their difference. Rows denote query sequences and columns denote database sequences; diagonal cells are excluded.}
    \label{fig:heatmap_zeroshot}
    \vspace{-1.0em}
\end{figure}

Fig.~\ref{fig:heatmap_zeroshot} decomposes the inter-sequence improvement into individual query--database pairs following the WildCross evaluation procedure~\cite{knights_wildcross_2026}. Most pairs improve with \coolname{}, with the largest Recall@1 gains occurring for K-04$\rightarrow$K-02 ($+13.1$), K-04$\rightarrow$K-03 ($+10.2$), V-04$\rightarrow$V-02 ($+9.4$), and K-02$\rightarrow$K-01 ($+9.1$). These gains concentrate on queries from the reverse-traversal and long-horizon appearance-variation sequences (sequences 02 and 04, respectively).
These results suggest that GFM-derived geometry is most useful where appearance alone is ambiguous, while the appearance representation remains broadly intact.

\begin{figure}
    \centering
    \includegraphics[draft=false,width=\columnwidth]{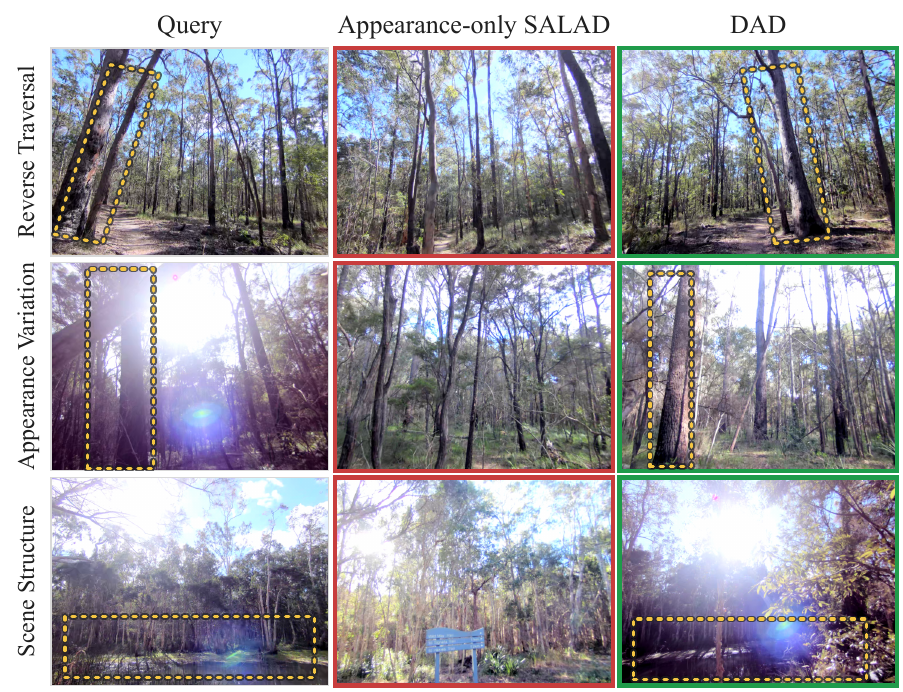}
     \caption{Representative WildCross retrievals illustrating reverse traversal, appearance variation, and distinctive scene structure. Each row shows a query, an incorrect top-ranked appearance-only SALAD retrieval, and the correct \coolname{} retrieval. Red and green borders denote incorrect and correct retrievals; dotted yellow outlines mark corresponding scene structure.}
    \label{fig:where-depth-helps}
    \vspace{-1.0em}
\end{figure}

\begin{figure}
    \vspace{-0.5em}
    \centering
    \includegraphics[draft=false,width=0.9\columnwidth]{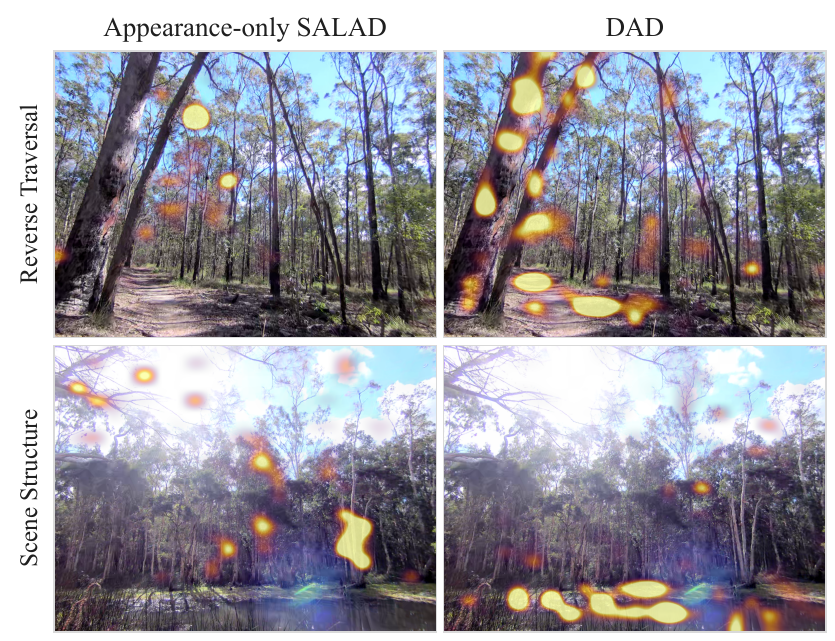}
    \caption{Local descriptor support for the reverse-traversal and scene-structure cases in Fig.~\ref{fig:where-depth-helps}. Columns compare the matched appearance-only SALAD baseline and \coolname{} on the same query image. Warmer regions indicate greater relative support within each independently normalized map.}
    \label{fig:descriptor-support}
    \vspace{-1.0em}
\end{figure}

The benefit of \coolname{} providing geometric grounding to the feature space is further reinforced in Fig.~\ref{fig:where-depth-helps}.
Here, we show representative examples of two challenging conditions and one case in which scene structure provides a distinctive cue.
In \textit{reverse traversal} (top row), the appearance-only model retrieves a path with superficially similar foliage, whereas \coolname{} returns the correct location despite having an opposing viewing angle thanks to features that focus on the structural components of the tree trunk shape and presence of the path. 
Under severe \textit{appearance variation} (middle row), glare and shadowing corrupt the RGB observation, while \coolname{} retrieves a target with a consistent forest layout. 
In the bottom-row scene-structure case, repeated texture and color provide weak discrimination, but the lake boundary and surrounding forest structure distinguish the correct location.

To demonstrate that the benefit of \coolname{} comes from the structural features used for matching images across sequences,  Fig.~\ref{fig:descriptor-support} shows a heatmap projection of local patch scores onto the original image, showing which patches contributed most to the final image-level descriptor used for final place recognition. 
Here we see that for a reverse-traversal example (top row) \coolname{} assigns strong support to the distinctive geometry of the foreground trunks and the path boundary, whereas the appearance-only system responds more sparsely to local canopy features. 
In the scene-structure case (bottom row), \coolname{} concentrates support along the partially occluded lake and its boundary, while the appearance-only descriptor responds primarily to glare, sky, and isolated foliage. 
These observations support the hypothesis that the geometry-conditioned representation places greater support on stable scene structure for place recognition.

\vspace{-1em}
\paragraph*{Ablations:}
\input{figures/DAD_ABLATIONS}

Table~\ref{tab:dad_ablations} isolates two central design choices in \coolname{}: staged optimization and channel-wise geometric gating. The appearance-only SALAD model is included as a matched control rather than an ablation. Descriptor distillation alone improves inter-sequence Recall@1 from 61.41 to 63.89, but remains below the control for intra-sequence retrieval, indicating that task refinement is required to obtain consistent gains across both settings. Optimising all \coolname{} components jointly from initialization instead lowers inter-sequence Recall@1 to 59.84, supporting the separation of descriptor distillation and task refinement. Finally, replacing the channel-wise gate $\boldsymbol{\alpha}\in\mathbb{R}^{C}$ with a scalar $\alpha\in\mathbb{R}$ reduces Recall@1 to 62.64 intra-sequence and 63.20 inter-sequence, supporting channel-specific regulation of the geometric contribution.

%% file: figures/DAD_RESULTS.tex
\begin{table*}[t!]
\vspace{-0.5mm}
\centering
\caption{Intra-sequence evaluation of fine-tuned VPR methods on each sequence of the \textit{WildCross} benchmark. Metrics are recall at 1 (R@1) and recall at 5 (R@5) with top result bolded. Our method (\coolname{}), highlighted in green is shown to generally improve results within intra-sequence setting.}
\label{tab:dad_intra_results}
\vspace{-0.5em}
\resizebox{\textwidth}{!}{%
\begin{tabular}{l|cccccccccccccccc|cc}
\toprule
\multirow{2}{*}{Method} &
\multicolumn{2}{c}{V-01} &
\multicolumn{2}{c}{V-02} &
\multicolumn{2}{c}{V-03} &
\multicolumn{2}{c}{V-04} &
\multicolumn{2}{c}{K-01} &
\multicolumn{2}{c}{K-02} &
\multicolumn{2}{c}{K-03} &
\multicolumn{2}{c|}{K-04} &
\multicolumn{2}{c}{Average} \\
& R@1 & R@5 & R@1 & R@5 & R@1 & R@5 & R@1 & R@5 & R@1 & R@5 & R@1 & R@5 & R@1 & R@5 & R@1 & R@5 & R@1 & R@5 \\
\midrule
NetVLAD~\cite{arandjelovic_netvlad_2016} & 68.95 & 72.16 & 69.89 & 74.05 & 18.09 & 22.72 & 59.90 & 66.02 & 83.86 & 87.53 & 86.32 & 91.57 & 26.82 & 32.35 & 56.32 & 59.96 & 58.77 & 63.30 \\
MixVPR~\cite{ali-bey_mixvpr_2023} & 66.16 & 68.52 & 68.86 & 72.78 & 17.80 & 25.01 & 52.90 & 57.34 & 84.66 & 87.89 & 87.99 & 89.59 & 30.40 & 38.22 & 60.87 & \textbf{65.04} & 58.70 & 63.05 \\
SALAD~\cite{izquierdo_optimal_2024} & 69.04 & 72.68 & 72.24 & 80.33 & 23.30 & 29.48 & 58.96 & 64.65 & 86.37 & 89.33 & 88.98 & 91.26 & \textbf{34.05} & 38.83 & \textbf{61.13} & 62.53 & 61.76 & 66.14 \\
\rowcolor{green!20}
\coolname{} (Ours) & \textbf{70.55} & \textbf{76.87} & \textbf{76.64} & \textbf{80.39} & \textbf{36.46} & \textbf{47.05} & \textbf{65.09} & \textbf{70.89} & \textbf{87.09} & \textbf{89.86} & \textbf{92.47} & \textbf{93.39} & 33.90 & \textbf{41.85} & 60.55 & 62.58 & \textbf{65.34} & \textbf{70.36} \\
\bottomrule
\end{tabular}
}
\vspace{-1.0em}
\end{table*}

\begin{table}[t]
\centering
\caption{Inter-sequence evaluation of fine-tuned VPR methods on \textit{WildCross}. Metrics are recall at 1 (R@1) and recall at 5 (R@5) with the top result in bold. Here we see the greatest benefit from our method (\coolname{}), in managing the visually ambiguous/inconsistent images across sequences.}
\label{tab:dad_inter_results}
\resizebox{\columnwidth}{!}{%
\begin{tabular}{l|cccc|cc}
\toprule
\multirow{2}{*}{Method} &
\multicolumn{2}{c}{Venman} &
\multicolumn{2}{c|}{Karawatha} &
\multicolumn{2}{c}{Average} \\
& R@1 & R@5 & R@1 & R@5 & R@1 & R@5 \\
\midrule
NetVLAD~\cite{arandjelovic_netvlad_2016} & 64.31 & 67.49 & 46.94 & 52.43 & 55.63 & 59.96 \\
MixVPR~\cite{ali-bey_mixvpr_2023} & 65.30 & 68.58 & 50.24 & 55.80 & 57.77 & 62.19 \\
SALAD~\cite{izquierdo_optimal_2024} & 68.54 & 71.86 & 54.29 & 59.86 & 61.41 & 65.86 \\
\rowcolor{green!20}
\coolname{} (Ours) & \textbf{72.17} & \textbf{76.83} & \textbf{60.57} & \textbf{68.15} & \textbf{66.37} & \textbf{72.49} \\
\bottomrule
\end{tabular}
}
\vspace{-1.0em}
\end{table}

%% file: figures/DAD_ABLATIONS.tex
\begin{table}[t!]
\vspace{-1.0em}
\centering
\caption{\coolname{} optimisation and gating ablations on \textit{WildCross}, reported as route-averaged Recall@1.}
\label{tab:dad_ablations}
\resizebox{\columnwidth}{!}{%
\begin{tabular}{lcc}
\toprule
Configuration & Intra avg. R@1 & Inter avg. R@1 \\
\midrule
\coolname{} (Ours) & \textbf{65.34} & \textbf{66.37} \\
\midrule
Appearance-only SALAD control & 61.76 & 61.41 \\
Descriptor Distillation only & 59.75 & 63.89 \\
Single-stage joint optimization & 61.11 & 59.84 \\
Scalar geometric gate ($\alpha\in\mathbb{R}$) & 62.64 & 63.20 \\
\bottomrule
\end{tabular}
}
\vspace{-1.0em}
\end{table}

%% file: sections/conclusion.tex
\section{Conclusion and Future Work}

We presented \coolname{}, a geometry-conditioned VPR framework for natural environments that incorporates GFM-derived scene structure into the token representation of a pretrained VFM without requiring a dedicated depth sensor. \coolname{} combines channel-wise geometric conditioning with a two-stage learning strategy: it first aligns the geometry-conditioned descriptor with an appearance-based teacher, then refines the resulting representation for place discrimination.
Experiments on WildCross show consistent improvements over the matched appearance-only VPR baseline, increasing average Recall@1 from 61.76\% to 65.34\% for intra-sequence retrieval and from 61.41\% to 66.37\% for inter-sequence retrieval. The largest improvements occur under reverse traversal and long-term appearance variation, where visual appearance becomes less reliable while scene structure remains comparatively persistent. These results demonstrate that GFM-derived geometry provides a complementary structural prior for improving VPR in challenging natural environments.